\documentclass[sigconf]{acmart} %
\makeatletter
\def\@ACM@checkaffil{
    \if@ACM@instpresent\else
    \ClassWarningNoLine{\@classname}{No institution present for an affiliation}%
    \fi
    \if@ACM@citypresent\else
    \ClassWarningNoLine{\@classname}{No city present for an affiliation}%
    \fi
    \if@ACM@countrypresent\else
        \ClassWarningNoLine{\@classname}{No country present for an affiliation}%
    \fi
}
\makeatother

\usepackage{algorithm}
\usepackage{algorithmicx}
\usepackage{algpseudocode}
\usepackage{multirow}
\usepackage{enumitem}
\usepackage{graphicx}
\usepackage{subfigure}
\usepackage{caption}
\usepackage{soul}
\usepackage{caption}
\usepackage{hyperref}
\usepackage{xspace}
\newcommand{\etal}{\emph{et al.}\xspace}
\newcommand{\eg}{\emph{e.g.,}\xspace}
\newcommand{\ie}{\emph{i.e.,}\xspace}
\newcommand{\etc}{\emph{etc.}\xspace}
\newcommand{\name}{HIEST\xspace}
\newcommand{\gcnname}{CHGCN\xspace}

\usepackage{xcolor}

\usepackage{color, xspace}

\author{Qian Ma}
\affiliation{%
  \institution{City University of Hong Kong}
  }
\email{qma233-c@my.cityu.edu.hk}
\authornote{Both authors contributed equally to this research.}

\author{Zijian Zhang}
\affiliation{%
  \institution{Jilin University}
  \institution{City University of Hong Kong}
   }
\email{zhangzj2114@mails.jlu.edu.cn}
\authornotemark[1]

\author{Xiangyu Zhao}
\affiliation{%
  \institution{City University of Hong Kong}
  }
\email{xianzhao@cityu.edu.hk}
\authornote{Xiangyu Zhao and Hongwei Zhao are the corresponding authors.}

\author{Haoliang Li}
\affiliation{%
  \institution{City University of Hong Kong}
  }
\email{haoliali@cityu.edu.hk}

\author{Hongwei Zhao}
\affiliation{%
  \institution{Jilin University}
  }
\email{zhaohw@jlu.edu.cn}
\authornotemark[2]

\author{Yiqi Wang}
\affiliation{%
  \institution{Michigan State University}
  }
\email{wangy206@msu.edu}

\author{Zitao Liu}
\affiliation{%
  \institution{
  Guangdong Institute of Smart Education, 
  Jinan University}
  }
\email{liuzitao@jnu.edu.cn}

\author{Wanyu Wang}
\affiliation{%
  \institution{City University of Hong Kong}
  }
\email{wanyuwang4-c@my.cityu.edu.hk}

\AtBeginDocument{%
  \providecommand\BibTeX{{%
    \normalfont B\kern-0.5em{\scshape i\kern-0.25em b}\kern-0.8em\TeX}}}

\copyrightyear{2023}
\acmYear{2023}
\setcopyright{acmlicensed}
\acmConference[CIKM '23] {Proceedings of the 32nd ACM International Conference on Information and Knowledge Management}{October 21--25, 2023}{Birmingham, United Kingdom.}
\acmBooktitle{Proceedings of the 32nd ACM International Conference on Information and Knowledge Management (CIKM '23), October 21--25, 2023, Birmingham, United Kingdom}
\acmPrice{15.00}
\acmISBN{979-8-4007-0124-5/23/10}
\acmDOI{10.1145/XXXXXX.XXXXXX}
\begin{document}
\renewcommand{\shortauthors}{Qian Ma et al.}
\copyrightyear{2023}
\acmYear{2023}
\setcopyright{acmlicensed}\acmConference[CIKM '23]{Proceedings of the 32nd
ACM International Conference on Information and Knowledge
Management}{October 21--25, 2023}{Birmingham, United Kingdom}
\acmBooktitle{Proceedings of the 32nd ACM International Conference on
Information and Knowledge Management (CIKM '23), October 21--25, 2023,
Birmingham, United Kingdom}
\acmPrice{15.00}
\acmDOI{10.1145/3583780.3614910}
\acmISBN{979-8-4007-0124-5/23/10}

\title{
Rethinking Sensors Modeling: Hierarchical Information Enhanced Traffic Forecasting\\
}

\begin{abstract}
With the acceleration of urbanization, traffic forecasting has become an essential role in smart city construction. In the context of spatio-temporal prediction, the key lies in how to model the dependencies of sensors. However, existing works basically only consider the micro relationships between sensors, where the sensors are treated equally, and their macroscopic dependencies are neglected. In this paper, we argue to rethink the sensor's dependency modeling from two hierarchies: regional and global perspectives. Particularly, we merge original sensors with high intra-region correlation as a region node to preserve the inter-region dependency. Then, we generate representative and common spatio-temporal patterns as global nodes to reflect a global dependency between sensors and provide auxiliary information for spatio-temporal dependency learning. In pursuit of the generality and reality of node representations, we incorporate a Meta GCN to calibrate the regional and global nodes in the physical data space. Furthermore, we devise the cross-hierarchy graph convolution to propagate information from different hierarchies. In a nutshell, we propose a Hierarchical Information Enhanced Spatio-Temporal prediction method, \name, to create and utilize the regional dependency and common spatio-temporal patterns. Extensive experiments have verified the leading performance of our \name against state-of-the-art baselines. We publicize the code to ease reproducibility\footnote{\url{https://github.com/VAN-QIAN/CIKM23-HIEST}}.

\end{abstract}



\maketitle
\section{Introduction}\label{sec:intro}

Intelligent Transportation Systems (ITS) play a critical role in smart city construction. Thanks to the promising progress of intelligent sensors, the collected extensive traffic data has unprecedentedly supported urban data mining research, such as traffic flow prediction \cite{flow1, flow2, flow3}, arriving time estimation \cite{time1, time2, autostl}, traffic speed analysis \cite{speed1, speed2, speed3, 10.1145/3580305.3599528} \etc
Spatio-temporal prediction aims to analyze historical spatio-temporal properties and predict future trends \cite{zhao2017modeling,zhao2022multi,zhao2022multi, zhao2018crime}. 
Inspired by the advancing capacity of representation learning of deep models, existing methods incorporate Temporal Convolution Networks (TCN) \cite{GWNET19,DMSTGCN}, Recurrent Neural Networks (RNN) \cite{DCRNN-17,tgcn}, and self-attention mechanism \cite{astgcn,gman} for temporal variation capture. Convolutional Neural Networks (CNN) \cite{ST-Res17,zhang2019flow,guo2019deep,li2020autost} and Graph Neural Networks (GNN) \cite{DCRNN-17,GWNET19,demystifying,source,jin2022automated} are generally utilized for spatial relationship learning.

\begin{figure*}
    \centering
    {\subfigure[{Original sensor graph and its regional graph. }]{\includegraphics[width=0.49\linewidth]{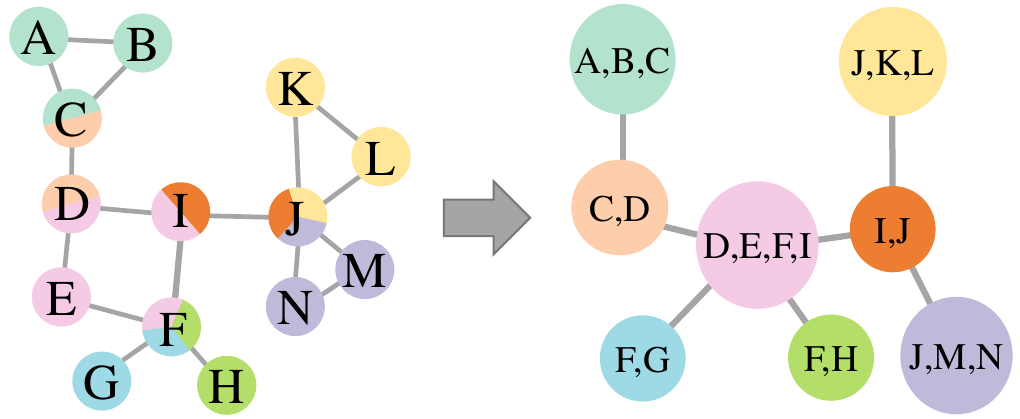}}}
    {\subfigure[{Regional graph and its global graph. }]{\includegraphics[width=0.49\linewidth]{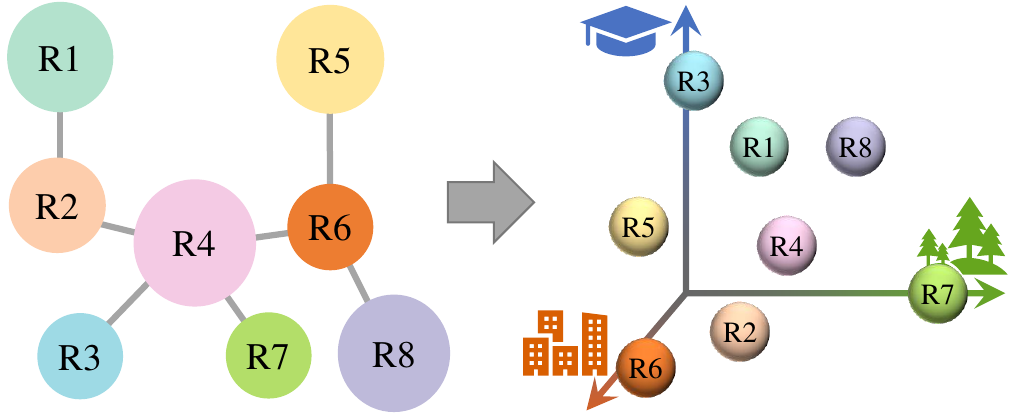}}}
    \caption{Illustration of the original sensor graph and two hierarchical perspectives. (a) Sensors with the same color belong to the same bi-connected component and can be merged from the regional perspective. Cut-vertices containing multiple colors are assigned to all corresponding regions. (b) Considering the properties of regions (such as functions), regions can be described by global nodes with common and representative spatio-temporal patterns as base vectors.}
    \label{fig:example}
\end{figure*}

To predict the traffic states of numerous sensors in a city, the key to solving spatio-temporal prediction is to capture the complex dependency among the sensors.
However, although having shown promising effectiveness, current solutions ~\cite{DCRNN-17,gman,gts,mtgnn, zhao2017modeling, zhao2017incorporating} 
of spatio-temporal prediction treat each sensor equally in sensor dependency modeling and essentially neglect the
dependencies among nonadjacent sensors
~\cite{hgcn21,traffictmr}. Sensors from different regions own similar or distinct properties, and considering sensors equally undermines the macro dependency learning.
In this paper, we argue to rethink the dependency modeling of different sensors.  We propose two hierarchical perspectives to depict the hierarchical dependencies among sensors, \ie regional and global perspectives.

\textbf{Regional perspective.} Sensors always share high-correlated properties with neighbors but weak relationships with others~\cite{hgcn21}. For example, sensors from the same county generally report similar traffic attributes since vehicles in the community flow in a relatively settled manner. Conversely, sensors maintain weak ties with the ones in other counties due to their unstable relationship.
Modeling the regional dependency properly can provide a hierarchical perspective to describe the sensor relationship and benefits the information propagation among sensors. 

\textbf{Global perspective.} Sensors own common spatio-temporal properties considering the function of their region~\cite{metast,HGNN20}. Traffic attributes recorded near schools should be different from those near natural parks because the weekday traffic in the educational areas varies regularly, but in recreation areas, the weekend traffic remarkably tops the weekdays. Moreover, different sensors from different educational areas are prone to have similar traffic attributes, \eg similar peaks and valleys of traffic flow variation in a day due to their similar traffic variation patterns. Hence, producing common and representative spatio-temporal patterns of sensors can offer auxiliary information to spatio-temporal modeling.


Basically, GCN-based methods \cite{GWNET19,DCRNN-17,stgcn,tgcn,mtgnn,gman} are representative solutions to model sensor relationships. DCRNN \cite{DCRNN-17} introduces diffusion convolution to propagate information in the graph. To dynamically describe the dependencies of sensors, GWNet \cite{GWNET19} generates adaptive adjacency matrices. 
Nevertheless, this series of works treats all fine-grained sensors equally regardless of the macro dependency among the sensors and their high-order neighbors.
Recently, efforts have been exerted to explore hierarchical sensor modeling \cite{HGNN20,22HiGCN,22ST-HSL,hgcn21,MCSTGCN22,traffictmr}. HGCN \cite{hgcn21} devises a hierarchical view based on the fine-grained sensors by simply clustering the node representations. To maintain the information from high-order neighbors, Traffic-Transformer \cite{traffictmr} resorts to masking the attention mechanism with high-order adjacency matrices. ST-HSL \cite{22ST-HSL} introduces a hyper-graph to address the macro relationship of sensors.
However, these methods build hierarchical views based on the node representations and demand tremendous computational overhead for clustering.
Besides, they neglect the important topological relationship, suffer from poor interpretability, and the performance depends on the manually designed architecture. 


To achieve the goals above, we identify several key research questions to be tackled. 
(i) \textit{How to formulate the macro dependency of sensors?}
(ii) \textit{How to attain the common and representative spatio-temporal patterns across the city?}
(iii) \textit{How to enhance spatio-temporal prediction given the macro dependency and representative spatio-temporal patterns of sensors?}
Conquering these challenges is far from trivial.
Given the numerous sensors with diverse properties: 
(i) Clustering the node representations lacks an understanding of graph structure, and the performance depends on the manually tuned architecture; 
(ii) Manual selection of representative sensors is impractical, and the generality and reality of spatio-temporal patterns are difficult to guarantee;
(iii) Most importantly, there is still a lack of a solution orchestrating the hierarchical information with the original sensor graph and consolidating the spatio-temporal dependency modeling effectively.

In this paper, we present a \textbf{H}ierarchical \textbf{I}nformation \textbf{E}nhanced \textbf{S}patio-\textbf{T}emporal prediction method, \textbf{\name}, to solve the research questions from the following three aspects.
\textit{Firstly, we propose to merge sensors with high local connectivity and achieve the graph of a regional perspective.}
Specifically,
instead of clustering nodes regardless of graph structure as in existing methods, we solve the Bi-Connected Component (BCC) of a graph, as in Figure \ref{fig:example} (a), and combine sensors belonging to the same BCC as the regional graph. We also assign the cut-vertex to the corresponding groups respectively, \eg node J is merged into all the related three groups, which further preserves the information propagation between sensors.
\textit{Based on the regional graph, we further extract general spatio-temporal patterns across the city with an orthogonal constraint to keep the distinctions.}
Through a meta graph convolution network, we generate representative sensors in the physical data space to maintain reality and generality.
These global nodes can serve as base vectors representing regions with diverse properties, as shown in Figure \ref{fig:example} (b).
Furthermore, \textit{we devise a Cross-Hierarchy Graph Convolutional Network, \gcnname, to enhance spatio-temporal prediction utilizing the merged sensors and general patterns.}
Our Meta GCN embeds the nodes in the original graph, regional graph, and global graph into the same feature space, to describe the spatio-temporal representation of the two hierarchical perspectives in the physical data space.
{Compared to existing hierarchical spatio-temporal prediction methods, the two hierarchies in \name owns physical meaning and good interpretability. 
We construct regional nodes according to the topological structure, and maintain the hierarchical nodes in the physical data space.}
Our main contributions are as follows:
\begin{itemize}[leftmargin=*]
    \item We reformulate spatio-temporal prediction with two hierarchical perspectives, \ie regional and global graphs, to capture the macro dependencies and common spatio-temporal patterns among sensors, respectively, which enjoy good interpretability. 
    \item We devise a cross-hierarchy graph representation learning schema to incorporate hierarchical graphs in sensors modeling and propose our model, \name. It maintains graph representations from the three perspectives in a meta-learning manner and keeps the hierarchical graph representations in the physical distribution.
    \item Extensive experimental verifications on three real-world datasets have proven the efficacy of our \name against state-of-the-art baselines. Comprehensive analyses have verified the performance of our proposed components.
\end{itemize}

\section{Preliminaries}
\noindent\textit{Definition 2.1.} \textbf{Spatio-Temporal Prediction.}\label{section:Formulation}
Let $G = (V, E)$ represent a sensor graph, 
where ${V}$ is the set of ${N}$ nodes and ${E}$ is the edge set.
Each node in the graph represents a traffic sensor in the road network.
The topological structure of $G$ is recorded by the adjacency matrix $\boldsymbol{A} \in  \{0, 1\}^{N \times N}$, where $\boldsymbol{A}_{i, j}=1$ denotes a connection between node $i$ and $j$.
Given the historical signal for $H$ timesteps, spatio-temporal prediction aims to predict the future $T$ timesteps graph signal:
\begin{equation}
    \boldsymbol{X}^{t-H:t}, G \xrightarrow{f_\theta} \boldsymbol{\hat{Y}}^{t+1:t+T}\label{eq:objective}
\end{equation}
\noindent where the $\boldsymbol{X}^{t-H:t} \in \mathbb{R}^{N \times D \times H} $ is historical graph signal and $\boldsymbol{\hat{Y}}^{t+1:t+T} \in \mathbb{R}^{N \times D \times T} $ is model prediction, $D$ is feature dimension.

\noindent\textit{Definition 2.2} \textbf{Connected Component (CC).}
If $G^{'}$ is a connected subgraph of $G$ and is not a part of any larger connected subgraph of $G$, then $G^{'}$ is a Connected Component~(CC) of~$G$.

\noindent\textit{Definition 2.3} \textbf{Cut Vertex.}
Node $v \in V$ is a cut vertex of $G$ if removing it will increase the number of CC in $G$.

\noindent\textit{Definition 2.4} \textbf{Bi-Connected Component (BCC).}\label{section:BBC}
If there is no cut vertex in graph $G$, then the graph $G$ will be considered as vertex-biconnected.
A vertex-biconnected component is regarded as a Bi-Connected Component (BCC)\footnote{In this paper, we denote Bi-Connected Component (BCC) as vertex-biconnected component \cite{BIC23}.}. 
For every two nodes from two different BCCs there exists one and only one path, \ie the path through the cut-vertex connecting the two BCCs.




\section{Framework}\label{sec:Framework}
\subsection{Framework Overview}

\begin{figure}[!t]
    \includegraphics[width=0.8\linewidth]{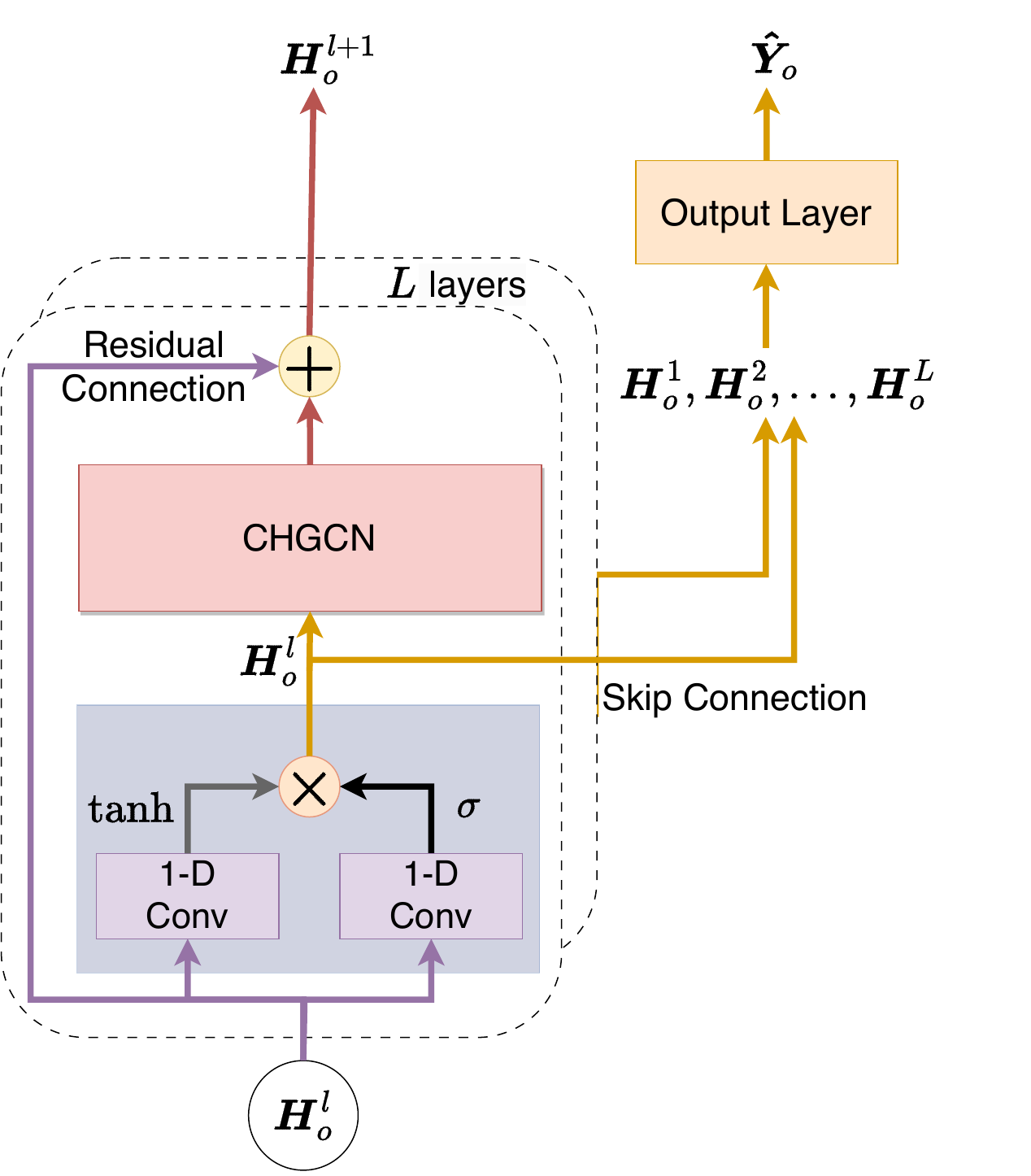}
    \caption{Framework overview of our \name.}\label{fig:framework}
\end{figure}

\begin{figure*}[!t]
    \includegraphics[width=0.8\linewidth]{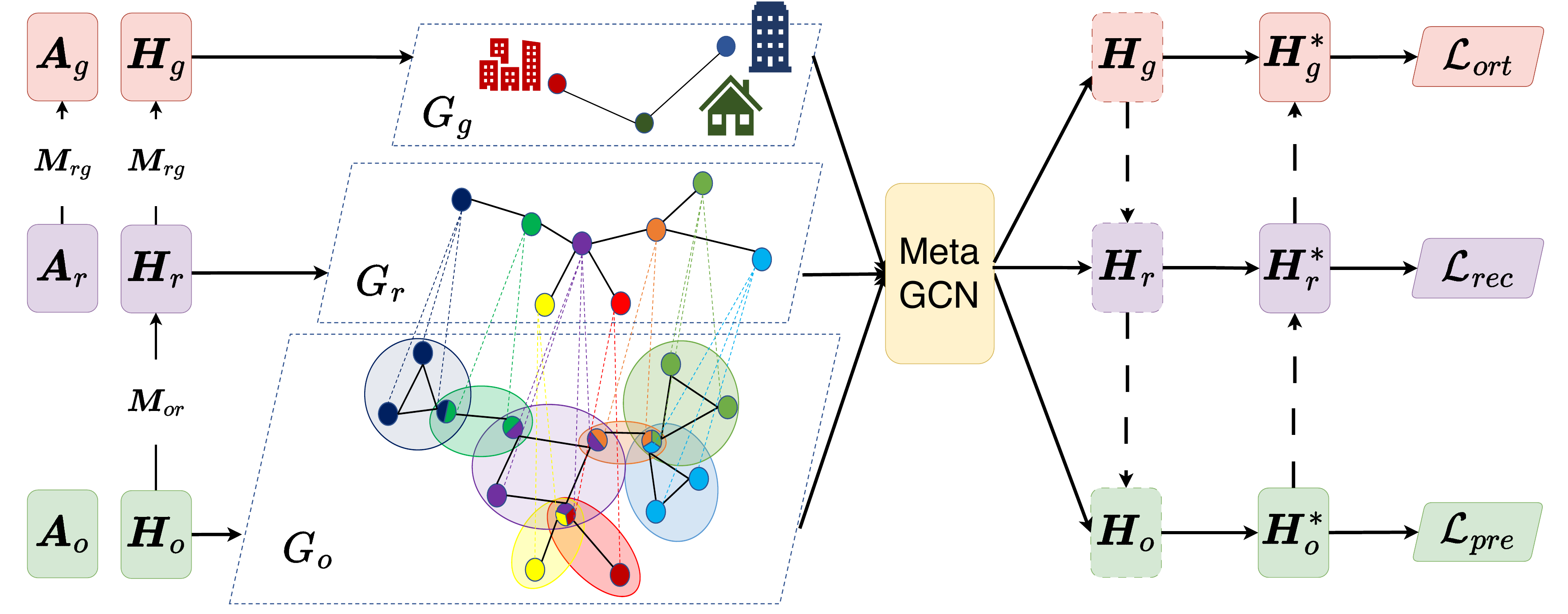}
    \caption{The pipeline of our \gcnname. Given $\boldsymbol{A}_o$, $\boldsymbol{A}_r$, and $\boldsymbol{H}_o$ as input, we generate $\boldsymbol{A}_g$ and $\boldsymbol{H}_g$ by $\boldsymbol{M}_{rg}$, and $\boldsymbol{H}_r$ by $\boldsymbol{M}_{or}$. Then we conduct graph convolution by Meta GCN, and update the node representation cross hierarchies.}\label{fig:THiGCN}
\end{figure*}
In this section, we detail our proposed framework, \name.
As shown in Figure \ref{fig:framework}, our \name mainly consists of $L$ layers of temporal convolution and graph convolution.
For temporal information capture, we introduce 1-D gated casual dilated convolution.
To model the complex relationship comprehensively, we elaborate a Cross-Hierarchy Graph Convolutional Network (\gcnname) to capture the proposed regional and global dependency among different nodes. 
After $L$ layers representation learning, we concatenate the intermediate representations of each layer and feed the output layer for final prediction. 
Table \ref{tab:notation} denotes the notations for a clear description.


Our \gcnname maintains graph convolution of the original graph, and two proposed hierarchical graphs, \ie regional and global graphs, as shown in Figure \ref{fig:THiGCN}, to capture our proposed regional and global dependencies. Moreover, we elaborate a cross-hierarchy information propagation to enhance the spatio-temporal dependency capture in each level and ultimately contribute to the representation learning of the original graph. 

\begin{table}[!t]
\small
\centering
\caption{Notation table.}\label{tab:notation}
\begin{tabular}{@{}lp{7cm}@{}}
\toprule
Notation & Definition                                                                                                  \\ \midrule
$N_o, N_r, N_g$        & Number of nodes in the original/regional/global graph                                                      \\
$\boldsymbol{A}_o,\boldsymbol{A}_r,\boldsymbol{A}_g$        & Adjacency matrix of the original/regional/global graph                                                     \\
$\boldsymbol{M}_{or},\boldsymbol{M}_{rg}$      & Mapping matrix to indicate the pairwise relationships between the original-regional and regional-global nodes \\
$\boldsymbol{H}_o,\boldsymbol{H}_r,\boldsymbol{H}_g$        & Feature matrix of the original/regional/global graph                                                       \\
\bottomrule
\end{tabular}
\vspace{-3mm}
\end{table}

\subsection{Regional Graph Construction}
\label{sec:regional_graph}
\subsubsection{Motivation.}
The original graph describes the pair-wise dependency between nodes.
Considering all nodes in an equivalent position in the original structure ignores the hierarchy of the graph. 
{In order to capture the regional perspective of nodes, we propose to merge nodes with close local dependencies into a region. 
Given the poor connectivity between nodes from different BCCs as in \textit{Definition 2.4} and the high local connectivity inside BCC, 
BCC describes the highly local dependencies in a graph and depicts its intrinsic structure \cite{BIC23}, which accords with our requirement.
}
As shown in Figure~\ref{fig:example} (a), node J is a cut-vertex and is the only entry across the BCCs (J,K,L) and (J,M,N). Traffic inside J, K, and L runs mutually, while traffic between J and M is relatively narrow due to the only connection J. So merging BCCs as regional nodes indicates the structure of the sensor graph and describes the macro dependency.

Existing works~\cite{MCSTGCN22, HGNN20,hgcn21} basically construct the regional nodes by spectral clustering, 
which do not utilize the topology information and can hardly reflect the skeleton of the road network. The cluster number is also manually assigned, which lacks interpretability. Besides, the hard mapping of clustering, \ie one node rigorously belongs to one cluster, also hinders the associations of nodes belonging to different clusters, which limits sensor dependency learning.


\subsubsection{Solving Bi-Connected Component}
To better group nodes with high local dependency, we propose a swift solution to describe the regional graph.
Specifically, we adapt Tarjan algorithm~\cite{tarjan}, a classical algorithm based on the Depth-First-Search approach, to solve the BCC given the graph. 
We construct a regional node in the regional graph $G_r$ by combining the nodes from the same BCC from the original graph $G_o$. 
The mapping matrix is constructed with the following steps.

Firstly, we establish the mapping matrix $\boldsymbol{M}_{or} \in \mathbb{R}^{N_o  \times N_r}$, which means the original sensor graph $G_o$ with $N_o$ nodes owns $N_r$ BCCs. 
$\boldsymbol{M}_{or}$ indicates if the original node $i$ is merged to the regional node $j$ according to the result of Tarjan algorithm.  
We initialize $\boldsymbol{M}_{or}[i, j]$ with $1$ to represent a mapping between original node $i$ and regional node $j$ and $0$ otherwise. 
Then we normalize $\boldsymbol{M}_{or} $
to keep the sum of each column of $\boldsymbol{M}_{or}$ equal to 1 and utilize it to generate the feature and adjacency matrix of the regional graph. 

In this context, given the feature matrix of the original graph $\boldsymbol{H}_o \in \mathbb{R}^{N_o  \times D} $, the feature matrix of the regional graph $\boldsymbol{H}_r \in \mathbb{R}^{N_r  \times D}$ will be aggregated according to the mapping matrix $\boldsymbol{M}_{or}$,
\begin{equation}
    \boldsymbol{H}_r = \boldsymbol{M}_{or}^T \cdot \boldsymbol{H}_o \label{eq:mapHor}
\end{equation}

Meanwhile, given the original graph $G_o$ with adjacency matrix $\boldsymbol{A_o} \in \mathbb{R}^{N_o \times N_o}$, the adjacency matrix of the regional graph $\boldsymbol{A}_{r} \in \mathbb{R}^{N_r\times N_r}$ will be generated according to the mapping matrix $\boldsymbol{M}_{or}$ to maintain the graph connectivity as follows,
\begin{equation}
    \boldsymbol{A}_r = \boldsymbol{M}_{or}^T \cdot \boldsymbol{A}_o \cdot\boldsymbol{M}_{or} \label{eq:mapAor}
\end{equation}

With the idea that original nodes with high dependency should be mapped together into regional nodes, we can use their representation to reconstruct the original graph.
To facilitate the graph representation learning of $G_r$, we deploy a binary cross-entropy function \cite{HGNN20} to compute the reconstruction loss $\mathcal{L}_{rec_{ro}}$ from the regional graph to the original graph $G_o$. 
We use the representation of the regional nodes to reconstruct the adjacency matrix of the original graph as follows,
\begin{align}
    & \boldsymbol{\hat{H}}_o = \boldsymbol{M}_{or}^T \cdot\boldsymbol{H}_r \label{eq:reconstruct0} \\ 
    & \boldsymbol{\hat{A}}_o = sigmoid(\boldsymbol{\hat{H}}_o \cdot \boldsymbol{\hat{H}}_o^T) \label{eq:reconstruct1} 
\end{align}
where the sigmoid function transforms the value between 0 and 1.

Then the reconstruction loss from regional to original graph $\mathcal{L}_{rec_{ro}}$ is constructed as follows,
\begin{equation}
    \begin{aligned}
    \mathcal{L}_{rec_{ro}} & = \frac{1}{N_o^2}\sum_{i,j \in V_o} - \boldsymbol{{A}}_o\left[i,j\right] \log \left(\boldsymbol{\hat{A}}_o[i,j]\right) \\
    & -\left(1-\boldsymbol{{A}}_o\left[i,j\right]\right)\log \left(1-\boldsymbol{\hat{A}}_o[i,j]\right) \label{eq:BCELossro}
    \end{aligned}
\end{equation}
where $V_o$ is the vertex set of the original graph $G_o$.

Through this combination, we can easily capture the high local dependency on topology information and avoid deciding the cluster number. Our solution achieves this goal effectively with linear time complexity of $O(|V|+|E|)$.
Furthermore, we repetitively assign the cut-vertex to the regional node corresponding to the BCC it belongs to, since it acts as a hub connecting multiple communities. The mapping matrix $\boldsymbol{M}_{or}$ is calculated offline and needs no update, which enjoys superior computational and maintenance costs.

\subsection{Global Graph Construction}
\subsubsection{Motivation.}
The regions with similar properties, such as functions or population density, usually have common spatial-temporal patterns that are different from those with different properties.
For example, the rush hour of the living areas is usually similar to each other, while different from the CBD areas.
Such common and representative spatio-temporal patterns across the city can offer auxiliary information to spatio-temporal representation learning of the original graph as a global dependency. 
The ideal situation is that some auxiliary data, \eg POI data, can be utilized to indicate the functionality of regions, \ie describes the global dependency explicitly. 
However, this is often not the case, and manually defining the common regions always lacks generality and demands tremendous expert effort. 
{To achieve the common spatio-temporal patterns across the city, we build a global view of the graph $G_g$, consisting of nodes with the most representative spatio-temporal patterns.}

\subsubsection{Generating Common Nodes.}
In this section, we generate common and representative spatio-temporal representations given the regional graph $G_r$. 
Basically, we utilize a trainable soft-mapping matrix to generate the global nodes.
Furthermore, we integrate the orthogonal constraint to pursue the global representations to be different from each other.

We set mapping matrix $\boldsymbol{M}_{rg} \in \mathbb{R}^{N_r\times N_g}$ to indicate the mapping between nodes in $G_r$ and $G_g$. It is noteworthy that $\boldsymbol{M}_{rg}$ is trainable according to model optimization.
Given the adjacency matrix of the regional graph $\boldsymbol{A}_r \in \mathbb{R}^{N_r \times N_r}$, the adjacency matrix of the global graph $\boldsymbol{A}_g \in \mathbb{R}^{N_g \times N_g}$ is generated as follows,
\begin{equation}
    \boldsymbol{A}_g = \boldsymbol{M}_{rg}^T \cdot \boldsymbol{A_r} \cdot \boldsymbol{M}_{rg} \label{eq:mapArg}
\end{equation}

Similarly, given the feature matrix of the regional graph $\boldsymbol{H}_r \in \mathbb{R}^{N_r \times D}$, the feature matrix of the global graph $\boldsymbol{X}_g \in \mathbb{R}^{N_g \times D}$ is generated as follows, 
\begin{equation}
    \boldsymbol{H}_g = \boldsymbol{M}_{rg}^T \cdot \boldsymbol{H}_r \label{eq:mapHrg}
\end{equation}

To keep representative spatio-temporal patterns in $G_g$, we integrate orthogonal loss to diversify the nodes' representations. {We minimize the cosine similarity of every two nodes in $G_g$,}
\begin{equation}
    \mathcal{L}_{ort}=\frac{1}{C_{N_g}^{2}} \sum_{i=1}^{N_g} \sum_{j=i+1}^{N_g} \left|\frac{{\boldsymbol{H}_g}[i] \odot \boldsymbol{H}_g[j]}{\left\|{\boldsymbol{H}_g}[i]\right\| \left\|\boldsymbol{H}_g[j]\right\|}\right| \label{eq:orthLoss}
\end{equation}
\noindent{where $C_{N_g}^{2}$ is the number of 2-combinations of $N_g$. }
Similarly, to enhance the graph representation learning, we incorporate reconstruction loss as in Equation~\eqref{eq:BCELossro}  to reconstruct the regional graph $G_r$ from the global graph $G_g$,
\begin{equation}
    \begin{aligned}
    \mathcal{L}_{rec_{gr}} & = \frac{1}{N_r^2}\sum_{i,j \in V_r} - \boldsymbol{{A}}_r\left[i,j\right] \log \left(\boldsymbol{\hat{A}}_r[i,j]\right) \\
    & -\left(1-\boldsymbol{{A}}_r\left[i,j\right]\right)\log \left(1-\boldsymbol{\hat{A}}_r[i,j]\right) \label{eq:BCELossgr}
    \end{aligned}
\end{equation}
where $V_r$ is the vertex set of the regional graph $G_r$.

It is worth notation that our trainable $\boldsymbol{M}_{rg}$ implements a soft mapping from regional nodes to global nodes, which is in line with the fact that each region can represent diverse properties.
Hence, our global nodes can serve as base vectors to describe any sensor considering the spatio-temporal patterns to contain the representative and common patterns of physical sensors.

\subsection{Hierarchical Graph Convolution}
\subsubsection{Meta Graph Convolution}
{To capture the spatial dependencies on the given graph and describe the nodes in the two hierarchical graphs in the physical data space, we introduce a Meta GCN to generate their graph signals.}
We incorporate diffusion convolution~\cite{DCRNN-17}, which is proven to be effective and calculated efficiently in spatial modeling, to capture the spatial dependency. 

We use a Meta GCN to process representation learning of $G_o$, $G_r$, and $G_g$, in order to describe nodes in three levels in the same data distribution.
By projecting the regional and global nodes to the data space of physical sensors, we can generate nodes of hierarchical levels with reality and generality.

Specifically, take the original graph $G_o$ as an example, given Graph $G_o$ and the adjacency matrix $\boldsymbol{A_o} \in \mathbb{R}^{N_o \times N_o}$ and input feature matrix $\boldsymbol{H_o} \in \mathbb{R}^{N_o \times D}$ of it, the general output $\boldsymbol{H_o} \in \mathbb{R}^{N_o \times D}$ of graph convolution~\cite{gcn} can be formulated as follows, 
\begin{equation}
    \boldsymbol{H_o} = \boldsymbol{A_o}\boldsymbol{H_o}\boldsymbol{W}
\end{equation} 
where $\boldsymbol{W} \in \mathbb{R}^{D \times D}$ is the trainable parameter matrix. The graph convolution is the same for $G_r$ and $G_g$.


\subsubsection{Hierarchical Information Enhanced Representation Learning}
With the three hierarchical graphs, we are able to maintain the dependency of nodes at three different levels. We propose hierarchical graph convolution to enhance representation learning with hierarchical graph information.
After processing by Meta GCN, we attain the updated representations of the three layers as $\boldsymbol{H}_o$, $\boldsymbol{H}_r$, and $\boldsymbol{H}_g$, respectively.
In order to strengthen the information propagation between different levels, we first enhance the node representation from top to bottom and then update backward, as shown in the dashed lines in Figure \ref{fig:THiGCN}.
Next, we introduce the cross-hierarchy representation learning of two steps, \ie the enhancement of $\boldsymbol{H}_o^*$ to integrate the information from common and regional hierarchies, and the representation update of $\boldsymbol{H}_r^*$ and $\boldsymbol{H}_g^*$.

To enhance the representation of the nodes in the original graph, 
we sequentially propagate the common information to the regional representation $\boldsymbol{H}_r$, and then the enhanced representation of the original graph $\boldsymbol{H}_o^*$ can be calculated as,
\begin{align}
    \boldsymbol{H}_r  & = \boldsymbol{H}_r +\eta_1 \cdot \sigma\left(\boldsymbol{M}_{r g} \cdot \boldsymbol{H}_g \right) \label{eq:gr} \\
    \boldsymbol{H}_o^{*} & = \boldsymbol{H}_o+\eta_2 \cdot \sigma\left(\boldsymbol{M}_{or} \cdot \boldsymbol{H}_r \right) \label{eq:ro}  
\end{align}

\noindent where $\sigma$ is the Relu activation function,
$\boldsymbol{M}_{or}$ and $\boldsymbol{M}_{rg}$ is the mapping matrix.
$\eta_1$ and $\eta_2$ are the hyperparameters to control the enhancement ratio. 

As the representation of the original graph is enhanced, we also update the representation of regional and global graphs in reverse.
We design the update process to propagate the node representation from the lower levels to higher levels and attain the node representation of regional graph $H_r^*$ and  global graph $H_g^*$, respectively,
\begin{align}
    \boldsymbol{H}_r^{*} & = \boldsymbol{H}_r+\eta_3 \cdot \sigma\left(\boldsymbol{M}_{or}^T \cdot \boldsymbol{H}_o^{*} \right) \label{eq:or} \\
    \boldsymbol{H}_g^{*} & = \boldsymbol{H}_g+\eta_4 \cdot \sigma\left(\boldsymbol{M}_{rg}^T \cdot \boldsymbol{H}_r^{*} \right) \label{eq:rg} 
\end{align}
where $\boldsymbol{M}_{or}^T$ and $\boldsymbol{M}_{r g}^T$ is the transverse of the mapping matrix $\boldsymbol{M}_{or}$ and $\boldsymbol{M}_{rg}$.
$\eta_3$ and $\eta_4$ are hyperparameters to control the update ratio.

\subsection{Temporal Dependency Learning}
For the temporal dependency learning, we incorporate the 1-D casual dilated convolution with gate mechanism, named TCN~\cite{TCN}.
Given the input feature matrix of the original graph $\boldsymbol{H_o} \in \mathbb{R}^{N  \times D}$, the output of the gated TCN can be presented as follows, 
\begin{equation}
    \boldsymbol{H_o} = tanh\left(\theta_1 \star \boldsymbol{H_o} + b_1 \right) \odot \sigma\left(\theta_2 \star \boldsymbol{H_o} + b_2\right) \label{eq:TCN}
\end{equation}
where $\sigma$ is the sigmoid function, $\odot$ is the element-wise production, $\theta_1$,  $\theta_2$, $b_1$, $b_2$ are the model parameters.

\subsection{Optimization}

The optimization objective function of our model contains four parts, the prediction loss $\mathcal{L}_{pre}$, the reconstruction loss from regional to original graph $ \mathcal{L}_{rec_{ro}}$, the reconstruction loss from global to regional graph $\mathcal{L}_{rec_{gr}}$ and the orthogonal loss $\mathcal{L}_{ort}$.
We deploy the MAE to evaluate the difference between the prediction and the ground truth as the prediction loss $\mathcal{L}_{pre}$ as follows,
\begin{equation}
    \mathcal{L}_{pre}=\frac{1}{T{N_o}} \sum_{i=1}^{T} \sum_{j=1}^{{N_o}} \left|{{\boldsymbol{\hat{Y}}}_j^i - \boldsymbol{Y}_j^i} \right| \label{eq:fineLoss}
\end{equation}

Hence the complete optimization objective function is,
\begin{equation}
    \mathcal{L}= \mathcal{L}_{pre} + \mathcal{L}_{rec_{ro}} + \mathcal{L}_{rec_{gr}} + \mathcal{L}_{ort}   \label{eq:objection}
\end{equation}

\section{Experiment}\label{sec:Experiments}

\subsection{Datasets}
The datasets used in this paper can be found in the public repos.
METR\_LA and PEMS\_BAY traffic speed datasets are released by Li \etal~\cite{DCRNN-17}
and PEMS\_D8 traffic flow datasets are released by Yan \etal~\cite{traffictmr}.
Each vertex on the graph represents a sensor to collect the traffic flow/speed data.
The summary statistics of the key elements of the datasets are shown in Table \ref{tab:dataset}.
Our graph construction follows the general setting~\cite{graph-construction}.
We constructed the adjacency matrix using distances between sensors with a threshold Gaussian kernel. We set the threshold to 0.1 for all datasets.

\subsection{Baselines}
We compare our proposed \name against the following baseline models on spatial-temporal prediction tasks.
{The baseline models can be divided into three main categories depending on how they process the given Graph. The first category considers the graph as a fixed, completed graph, such as DCRNN ~\cite{DCRNN-17}, GMAN~\cite{gman}, STGCN~\cite{stgcn}, and TGCN ~\cite{tgcn}. The second category considers the graph as incomplete and learns potential links, such as GWnet~\cite{GWNET19}, GTS~\cite{gts}, and MTGNN~\cite{mtgnn}. The third category aims to extract hierarchical information from the given graph, including HGCN~\cite{HGNN20} and Trans~\cite{traffictmr}.}
\begin{itemize}[leftmargin=*] 
\item SVR: A ML regression method for time series ~prediction.

\item DCRNN: Diffusion graph convolutional networks applied as the spatial feature extractor and RNN as the temporal feature extractor to extract spatial-temporal correlations.

\item STGCN: STGCN integrates graph convolution and gated time convolution to capture spatio-temporal correlations.

\item GraphWavenet (GWnet):  Casual Dialated Convolution for extracting the temporal features. Diffusion convolution is adapted for the graph convolution layer of GraphWavenet.

\item TGCN: TGCN combines the GCN and GRU for spatial-temporal predicting.

\item GMAN: GMAN applies attention mechanism to extract spatial and temporal features for prediction.

\item MTGNN: MTGNN can model potential spatial dependencies between pairs of variables in multivariate time series data.

\item GTS: GTS proposes learning the graph structure simultaneously with the GNN if the graph is unknown. 

\item HGCN: HGCN is a hierarchical-based ST prediction model with spectral clustering. 

\item Traffic-Transformer (Trans): Trans proposes a hierarchical-based Transformer with a global encoder and global–local decoder to extract and fuse the spatial-temporal patterns globally and locally.

\end{itemize}

\begin{table}[!t]
\small
\centering
\caption{Datasets statistics. }\label{tab:dataset}
    \begin{tabular}{@{}ccccc@{}}
    \toprule
    Dataset  & \#Nodes & \#Edges & \#Timesteps & Period                  \\ \midrule
    PEMS\_BAY & 325     & 2,369   & 34,272      & 2017/01/01 - 2017/06/30    \\
    PEMS\_D8  & 170     & 276     & 17,856      & 2016/07/01 - 2016/08/31 \\
    METR\_LA  & 207     & 1,515   & 52,116      & 2012/03/01 - 2012/06/27 \\ \bottomrule
    \end{tabular}
\end{table}

\begin{table*}[!t]
\small
\centering
\caption{Overall performance comparison. Best performances are bold, next best are underlined.}
\label{tab:overall}
\begin{tabular}{p{2cm}lcccccccccc}
\cmidrule(r){1-12}
\multicolumn{1}{l}{\multirow{2}{*}{Dataset}} & \multirow{2}{*}{Model}  & \multicolumn{3}{c}{15 min} & \multicolumn{3}{c}{30 min} & \multicolumn{3}{c}{60 min} & \multirow{2}{*}{Params(K)} \\ \cmidrule(lr){3-5}\cmidrule(lr){6-8}\cmidrule(lr){9-11} 
\multicolumn{1}{l}{}                         &                         & MAE     & MAPE    & RMSE   & MAE     & MAPE    & RMSE   & MAE     & MAPE    & RMSE   & \\ \cmidrule(l){1-12}
\multirow{11}{*}{METR\_LA}                    
                                             & SVR   & 4.23    & 9.46\%  & 9.97   & 5.28    & 11.71\% & 12.27  & 7.00    & 15.59\% & 15.33  & -\\
                                             & DCRNN & 2.55    & 6.54\%  & 5.08   & 3.02    & 8.27\%  & 6.29   & 3.57    & 10.40\%  & 7.60 & 372 \\
                                             & STGCN & 2.76    & 7.13\%  & 5.10   & 3.43    & 9.61\%  & 6.60   & 4.31    & 12.62\%  & 8.26  & 320\\
                                             & GWnet & 2.55    & 6.67\%  & 5.19   & 2.96    & 8.11\%  & 6.26   & 3.39    & \underline{9.57\%}  & 7.29  & 276\\                            
                                             & TGCN  & 3.11    & 8.22\%  & 5.42   & 3.63    & 9.97\%  & 6.39   & 4.30    & 12.17\%  & 7.46  & 32\\
                                             & GMAN  & 2.85    & 7.56\%  & 5.17   & 3.27    & 9.08\%  & \underline{6.10}   & 3.72    & 10.68\%  & \textbf{7.00}  & 900\\
                                             & MTGNN & 2.49    & \underline{6.31\%}  & 5.03   & \underline{2.91}    & \underline{7.91\%}  & 6.12   & \underline{3.38}    & \textbf{9.54\%}  & {7.24}  & 405 \\                                            
                                             & GTS   & \underline{2.48}    & 6.35\%  & \underline{5.01}   & \underline{2.91}    & 7.96\%  & 6.13   & 3.40    & 9.90\%  & 7.35   & 38,494\\
                                             & HGCN  & 2.59    & 6.70\%  & 5.17   & 3.06    & 8.45\%  & 6.35   & 3.60    & 10.40\%  & 7.55  & 690\\
                                             & Trans  & 2.68    & 6.97\%  & 5.33   & 3.11    & 8.54\%  & 6.50   & 3.58    & 10.13\%  & 7.62  & 220\\
                                             
                                             & \textbf{\name}& \textbf{2.46}    & \textbf{6.19\%}  & \textbf{4.91}   & \textbf{2.89}    & \textbf{7.76\%}  & \textbf{6.02}   & \textbf{3.37}    & \textbf{9.54\%}  & \underline{7.16} & 275  \\ \cmidrule(r){1-12}\cmidrule(lr){3-12}
\multirow{11}{*}{PEMS\_BAY}                  & SVR   & 1.61    & 3.87\%  & 3.58   & 2.00    & 4.76\%  & 4.58   & 2.69    & 6.44\%  & 6.20  & -\\
                                             & DCRNN & \underline{1.32}    & 2.74\%  & \underline{2.78}   & 1.66    & 3.70\%  & 3.79   & 2.07    & 4.64\%  & 4.64  & 372 \\
                                             & STGCN & 1.38    & 2.89\%  & 2.83   & 1.82    & 4.07\%  & 3.99   & 2.37    & 5.32\%  & 5.12  & 320 \\
                                             & GWnet & \underline{1.32}    & 2.77\%  & 2.79   & 1.65    & 3.71\%  & 3.74   & 1.98    & 4.59\%  & 4.54  & 279 \\
                                             & TGCN  & 1.60    & 3.36\%  & 3.02   & 2.00    & 4.44\%  & 3.99   & 2.53    & 5.79\%  & 4.95  & 32 \\
                                             & GMAN  & 1.51    & 3.21\%  & 2.97   & 1.82    & 4.06\%  & 3.86   & 2.10    & 4.79\%  & 4.53  & 900 \\
                                             & MTGNN & \underline{1.32}    & 2.77\%  & \underline{2.78}   & 1.64    & 3.64\%  & \underline{3.71}   & 1.94    & \textbf{4.41\%}  & 4.47  & 573 \\
                                             
                                             & GTS   & \textbf{1.30}    & \underline{2.70}\%  & \underline{2.78}   & \textbf{1.61}    & \textbf{3.58\%}  & \underline{3.71}   & \textbf{1.90}    & \underline{4.43\%}  & \underline{4.45} &58,462  \\
                                             & HGCN  & 1.35    & 2.88\%  & 2.85   & 1.70    & 3.91\%  & 3.85   & 2.03    & 4.84\%  & 4.62  & 766 \\
                                             & Trans  & 1.40    & 2.96\%  & 2.89   & 1.71    & 3.77\%  & 3.81   & 2.04    & 4.58\%  & 4.65  & 220\\
                                              
                                             & \textbf{\name}& \textbf{1.30}    & \textbf{2.69\%}  & \textbf{2.74}   & \underline{1.63}    & \underline{3.63\%}  & \textbf{3.68}   & \underline{1.93}    & 4.50\%  & \textbf{4.44} & 275 \\\cmidrule(r){1-12}\cmidrule(lr){3-12}
\multirow{11}{*}{PEMS\_D8}                   & SVR   & 93.08   & 66.81  & 120.29   & 93.01    & 67.12  & 120.36   & 92.92    & 67.78  & 120.59  & -\\
                                             & DCRNN & 15.13   & 9.82\%  & 23.48  & 16.66   & 10.79\% & 26.09  & 19.43   & 12.51\% & 30.32 & 372 \\
                                             & STGCN & 14.95   & 9.63\%  & 23.24  & 16.85   & 10.87\%  & 26.13   & 20.17    & 12.84\%  & 30.91  & 296\\
                                             & GWnet & 14.49   & 9.66\%  & 22.89  & 15.76   & 11.05\% & 25.14  & 18.17   & 12.46\% & 28.84 & 289 \\
                                             & TGCN  & 16.68   & 12.82\% & 25.18  & 18.00   & 13.70\%  & 27.28   & 21.19    &16.58\%  & 31.67  & 32\\
                                             
                                             & GMAN  & 14.37   & 10.27\% & 22.85  & 15.23   & 10.75\% & 24.51   & 17.00    & 12.09\%  & 27.02   & 900 \\
                                             & MTGNN & 13.94   & 9.02\%  & 22.04  & 14.84   & \underline{9.53\%}  & {23.76}   & 16.65  & 10.90\%  & \underline{26.35}  & 352 \\
                                             & GTS   & \underline{13.62}   & \underline{8.81\%}  & \underline{21.84}  & \underline{14.59}   & \underline{9.53\%}   & \underline{23.68}   & \underline{16.26}    & \underline{10.79\%}  & 26.43 & 20,164  \\
                                             & HGCN  & 15.44   & 10.58\% & 26.30  & 16.91   & 11.88\%  & 26.39   & 19.87    &14.97\%  & 30.43  & 671\\
                                             & Trans  & 17.78    & 13.02\%  & 26.94  & 19.02   & 13.89\% & 28.98   & 22.34   & 16.10\%  & 33.93  & 220\\
                                             
                                             & \textbf{\name}& \textbf{13.42}   & \textbf{8.62\%}  & \textbf{21.66}  & \textbf{14.23}   & \textbf{9.31\%}  & \textbf{23.50}	& \textbf{15.64} & \textbf{10.23\%}  & \textbf{25.97} & 275\\  \cmidrule(r){1-12}                                             
\end{tabular}
\end{table*}

\subsection{Experiments Setups}
We split the data into training, validation, and testing sets with a ratio of 7:1:2 for all datasets. The hidden size is 32. The number of blocks and layers is 4 and 2, respectively.
We predict the spatio-temporal attributes \ie traffic speed or flow, of the future 12 timesteps based on the historical 12 timesteps, \ie $T$ = $H$ = 12. 
For the hyperparameters of our proposed \name, we set $\eta1,\eta2,\eta3,\eta4 = 1 $ and $N_g = 15 $ for all datasets.

To comprehensively evaluate the methods, we present the performance of mean absolute error~\textbf{MAE}, mean absolute percentage error~\textbf{MAPE}, and root mean squared error~\textbf{RMSE}.
To ensure a fair comparison between our \name and the baselines, we refer to the public repository LibCity~\cite{libcity}. All experiments are executed on 1 NVIDIA V100s with the random seed set to 0 and repeated thrice. 

\subsection{Overall Performance}
The overall results over 3 real-world datasets are shown in Table~\ref{tab:overall}.
From the performance comparison against state-of-the-art baselines, several conclusions can be made:


\name consistently outperforms DCRNN, GMAN, STGCN, and TGCN. These baseline models consider the sensors equally without any hierarchical information capture. In contrast, our proposed \name uses a regional and global graph to describe the macro-structure of the original sensor graph and capture implicit higher-order neighborhood relations. This enables \gcnname to handle spatio-temporal representation learning comprehensively.
 
GWnet, GTS, and MTGNN propose to model the intrinsic relationship among nodes by establishing trainable adjacency matrices or optimizing a probabilistic graph model.
Dense data with plenty of sensors can foster their proposed intrinsic dependency modeling.
Our proposed method achieves competitive performance against them with only 48\% of the parameters used by MTGNN and 0.5\%-0.7\% of GTS. On the PEMS\_D8 dataset with fewer nodes, edges, and recordings, our proposed \name outperforms them by utilizing the proposed global graph nodes, whose universal and representative spatio-temporal representation is guaranteed by orthogonal constraints of \gcnname.

Next, we discuss the two hierarchical spatio-temporal prediction baselines. 
\name outperforms HGCN on all settings. HGCN conducts a new hierarchy above the original graph by spectral clustering, while our proposed method conducts new hierarchies from two perspectives: the regional and global graphs described in the real data space. 
The better results of \name over HGCN demonstrate the superiority of our proposed two new hierarchies over the simple hierarchy in HGCN.

\name also outperforms Trans on all three datasets. Trans proposes a global encoder with a multi-head attention block to capture common global spatial-temporal patterns, ignoring the topological structure of the graph. 
It masks the attention matrix based on the high-order adjacency matrix to incorporate hierarchical neighbor information.
In contrast, our solver for BCCs requires linear computational complexity and enjoys promising interpretability considering the topology structure.

\name achieves competitive performance against the state-of-the-art baselines across all horizons, which demonstrates its ability for both short-term and long-term prediction. 
Its two hierarchical graphs capture the macro dependency and common properties of sensors, and the cross-hierarchy graph convolution network well enhances the information propagation among different levels, then contributes to the spatio-temporal depedency learning of sensors.

\subsection{Ablation Study}
{The two most critical components of our \name are the two new hierarchies, \ie regional and global perspectives, and the cross-hierarchy representation learning consisting of enhance process and update process. To verify their effectiveness explicitly, we generated the following variations of our model:}
\begin{itemize}[leftmargin=*]
    \item $w/o\ EU$: \name without the enhance and update process across all three layers.
    \item $w/o\ U$:  \name without the update of $\boldsymbol{H}_r^*$ and $\boldsymbol{H}_g^*$.
    \item $w/o\ G_g$: \name without the global graph, \ie consisting of original graph $G_o$ and $G_r$.
    \item $w/o\ G_r$: \name without the regional graph, \ie consisting of original graph $G_o$ and $G_g$.
\end{itemize}




\begin{table}
\small
\centering
\setlength{\tabcolsep}{2.5pt} 
\caption{Ablation study results.}
\label{tab:Ablation}
\begin{tabular}{p{1cm}lcccccccccc}
\cmidrule(r){1-10}
\multirow{2}{*}{Model}  & \multicolumn{3}{c}{15 min} & \multicolumn{3}{c}{30 min} & \multicolumn{3}{c}{60 min} \\ \cmidrule(lr){2-4}\cmidrule(lr){5-7}\cmidrule(lr){8-10} 
            & MAE     & MAPE    & RMSE   & MAE     & MAPE    & RMSE   & MAE     & MAPE    & RMSE    \\ \cmidrule(r){1-10}
$w/o\ EU$   & 14.31   & 9.09\%  & 22.84  & 15.57   & 10.02\%  & 25.20   & 17.79    & 11.51\% & 28.89 \\
$w/o\ U$  & \underline{13.73}   & \underline{9.05\%}  & \underline{22.17}  & \underline{14.68}   & 9.66\%   & \underline{23.97}   & \underline{16.32}    & \underline{10.59\%} & \underline{26.47} \\
$w/o\ G_g$    & 14.02   & 9.23\%  & 22.41  & 14.90   & \underline{9.57\%}   & 24.32   & 16.80    & 11.08\% & 27.54  \\
$w/o\ G_r$ & 14.95   & 9.63\%  & 23.24  & 16.85   & 10.87\%  & 26.13   & 20.17    & 12.84\% & 30.91   \\
\textbf{\name}      & \textbf{13.42}   & \textbf{8.62\%}  & \textbf{21.66}  & \textbf{14.23}   & \textbf{9.31\%}  & \textbf{23.50}	& \textbf{15.64} & \textbf{10.23\%}  & \textbf{25.97}  \\  \cmidrule(r){1-10}                                             
\end{tabular}
\end{table}

Table~\ref{tab:Ablation} presents the results of the ablation study.
We can draw a conclusion that both the two proposed hierarchies and the enhanced information are critical to our proposed model. 
The superiority of \name over $w/o\ EU$ proves that our proposed hierarchical update mechanism and generic nodes can foster the learning of the spatio-temporal representation of the original graph.
The results of $w/o\ U$ demonstrate that the information propagation from lower layers to higher ones can also benefit spatio-temporal dependency capture, which emphasizes the efficacy of our cross-hierarchy representation learning. 
The results of $w/o\ G_g$ indicate that our generated common representation offers essential information to the original sensor graph modeling.  
$w/o\ G_r$ achieves poor results without the regional graph describing the macro dependency. The global graph can not be well constructed directly over the sensor-level, which proves the validity of our two progressively hierarchical perspectives.

\subsection{{Hyperparameter Analysis}}
To further verify the efficacy of our proposed global graph and the update process from the regional graph to the global graph, with $\eta1,\eta2,\eta3 = 1 $, we conduct the analysis for the hyperparameters, $N_g \in \{10,15,20,25\}$ and $\eta4 \in \{0,0.2,0.5,0.8,1\}$, of our proposed model \name on the PEMS\_D8 datasets: 
\begin{table}
\vspace{-2mm}
\small
\centering
\setlength{\tabcolsep}{3pt} 
\caption{Parameter analysis results for $\eta4$.}
\label{tab:eta}
\begin{tabular}{p{0.3cm}ccccccccccc}
\cmidrule(r){1-10}
\multirow{2}{*}{$\eta4$}  & \multicolumn{3}{c}{15 min} & \multicolumn{3}{c}{30 min} & \multicolumn{3}{c}{60 min}  \\ \cmidrule(lr){2-4}\cmidrule(lr){5-7}\cmidrule(lr){8-10} 
         & MAE     & MAPE    & RMSE   & MAE     & MAPE    & RMSE    & MAE     & MAPE    & RMSE  \\ \cmidrule(r){1-10}
0        & \underline{13.63}   & 9.04\%  & \underline{21.88}  & 14.50   & 9.48\%  & \underline{23.63}   & 16.08   & 10.46\%  & 26.27 \\
0.2      & 13.65   & 8.94\%  & 21.92  & 14.52   & 9.52\%  & 23.76   & 16.05   & 10.49\%  & 26.23 \\
0.5      & 13.64   & 8.84\%  & \underline{21.88}  & \underline{14.49}   & 9.45\%  & 23.71   & \underline{15.94}   & \underline{10.39\%}  & \underline{26.17} \\
0.8      & 13.71   & \underline{8.80\%}  & 22.01  & 14.66   & \underline{9.40\%}  & 24.09   & 16.24   & 10.62\%  & 26.61 \\
1        & \textbf{13.42}   & \textbf{8.62\%}  & \textbf{21.66}  & \textbf{14.23}   & \textbf{9.31\%}  & \textbf{23.50}	& \textbf{15.64} & \textbf{10.23\%}  & \textbf{25.97} \\  \cmidrule(r){1-10}                                             
\end{tabular}
\end{table}

Table~\ref{tab:eta} presents the results of applying different update ratios $\eta_4$, and those less than 1 indicate that the updated regional representation should be given equal importance to provide comprehensive information when updating the global representation. Otherwise, it can be considered as noise that contradicts the good mapping between the regional and global graphs.
\begin{table}
\small
\vspace{-2mm}
\centering
\setlength{\tabcolsep}{3pt} 
\caption{Parameter analysis results for $N_g$.}
\label{tab:NG}
\begin{tabular}{p{0.3cm}ccccccccccc}
\cmidrule(r){1-10}
\multirow{2}{*}{$N_g$} & \multicolumn{3}{c}{15 min} & \multicolumn{3}{c}{30 min} & \multicolumn{3}{c}{60 min} \\ \cmidrule(lr){2-4}\cmidrule(lr){5-7}\cmidrule(lr){8-10}
                    & MAE     & MAPE    & RMSE  & MAE   & MAPE   & RMSE  & MAE   & MAPE   & RMSE  \\ \cmidrule(r){1-10}
10                  & 13.54   & \underline{8.76\%}  & 21.79 & 14.43 & \underline{9.44\%} & {23.62} & 15.96 & 10.57\% & 26.21      \\
15                  & \textbf{13.42}   & \textbf{8.62\%}  & \textbf{21.66}  & \textbf{14.23}   & \textbf{9.31\%}  & \underline{23.50}	& \underline{15.64} & \underline{10.23\%}  & \underline{25.97}  \\
20                  & {13.50}   & 9.14\%  & {21.77}  & {14.42}   & 9.97\% & 23.67   & {15.82}   & {10.38\%}  & {26.15}   \\ 
25                  & \underline{13.47}   & 9.04\%  & \underline{21.74}  & \underline{14.26}   & 9.49\% & \textbf{23.47}   & \textbf{15.59}   & \textbf{10.08\%}  & \textbf{25.72}  \\ \cmidrule(r){1-10}    
\end{tabular}
\end{table}

Table~\ref{tab:NG} shows that different numbers of global nodes $N_g$ lead to comparable results, which are generally better than baselines. 
Our proposed model is not very sensitive to the number of global nodes, and both dense or sparse base vectors can well describe the vector space of sensors spatio-temporal patterns.

\subsection{Visualization}
\begin{figure*}[htbp]
{\subfigure[{The mapping results from the original to regional graph }]{\includegraphics[width=0.49\linewidth]{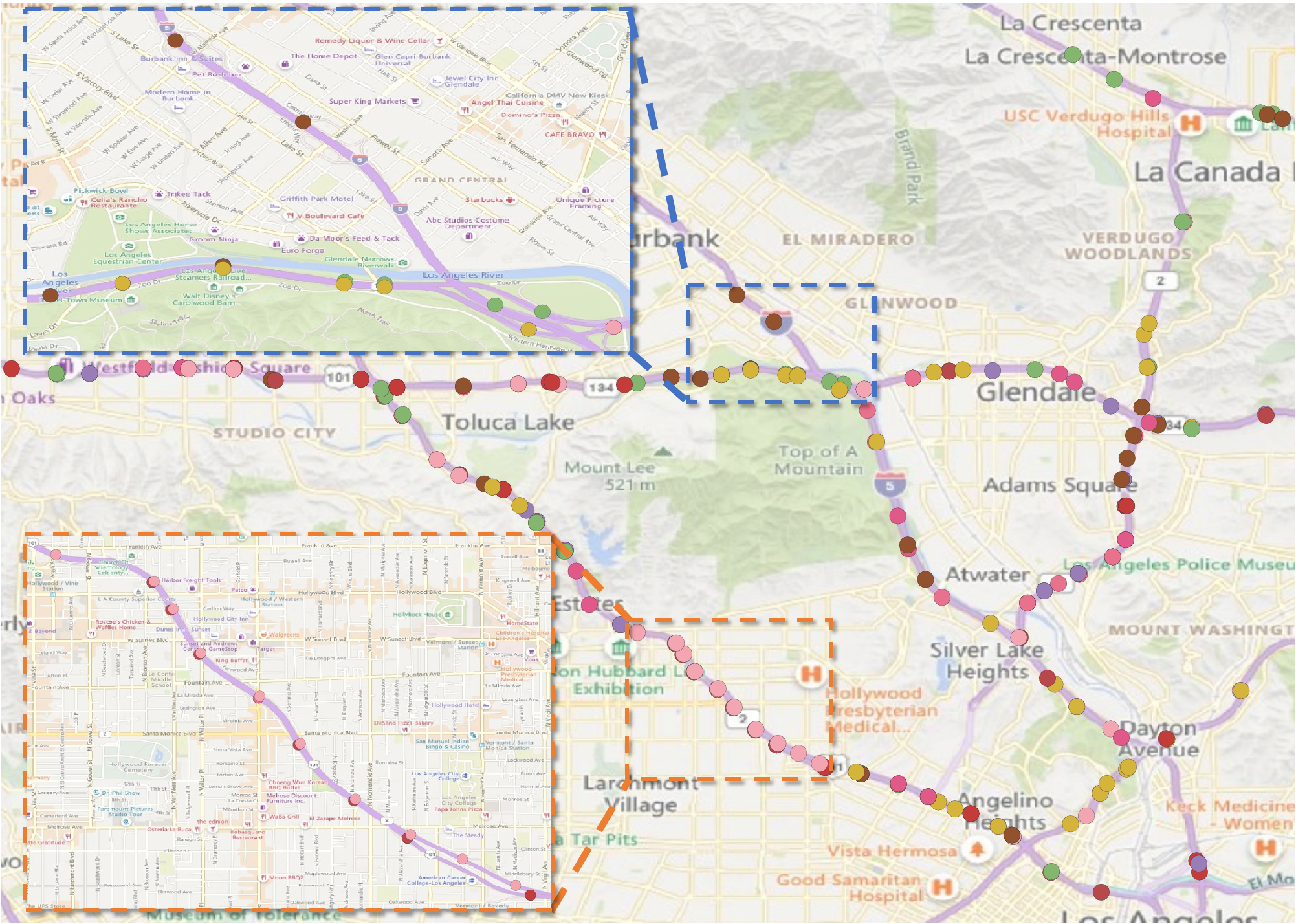}}}
{\subfigure[{The mapping results from the original to global graph }]{\includegraphics[width=0.49\linewidth]{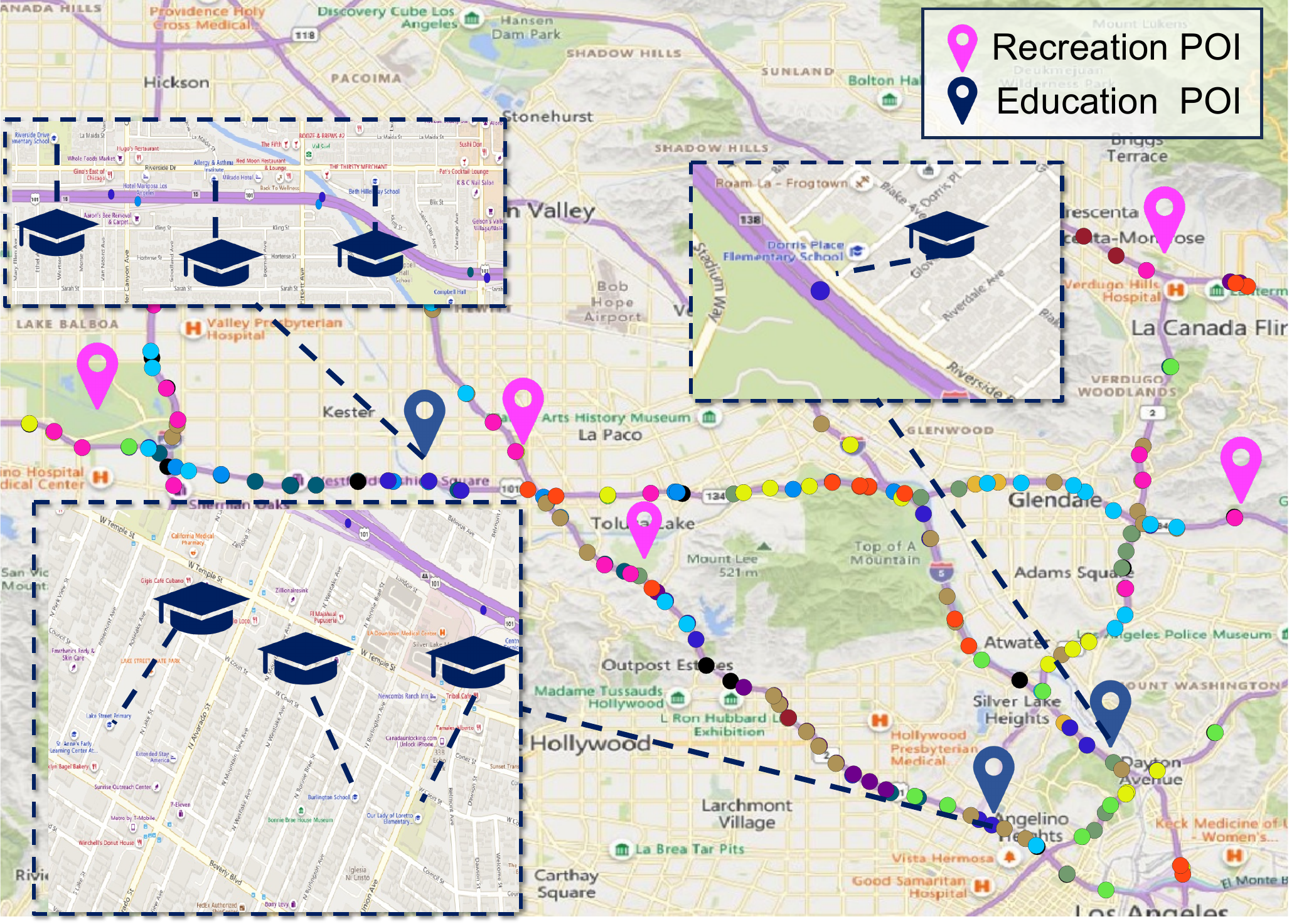}}}
    \vspace*{-1mm}
    \caption{The visualization on METR\_LA.}\label{fig:METR_LA_Mog}
\end{figure*}

To achieve a close view of our \name, we visualize the two hierarchical graphs $G_r$ and $G_g$ on METR\_LA, as shown in Figure \ref{fig:METR_LA_Mog}.
Specifically, we color original nodes with their related regions as Figure \ref{fig:METR_LA_Mog} (a). The 207 sensors in $G_o$ are divided into 136 regions in $G_r$, and we color using 10 colors according to the constraint that any two adjacent nodes from different regions are within different colors.
For Figure \ref{fig:METR_LA_Mog} (b), we illustrate the common and representative properties of regions and mark the POIs near sensors. 
We generate 15 global nodes, \ie 15 colors, to depict which global node the sensor belongs to.
Notably, considering that each region is assigned a weight vector of all global nodes in the soft-mapping matrix $\boldsymbol{M}_{rg}$, we choose the global node with the highest weight for illustration.

For the mapping results from the regional perspective in Figure \ref{fig:METR_LA_Mog} (a), it is in accordance with the expectation that the regional node depicts region properties to some extent.
We zoom in on two areas of the blue and orange rectangles.
In the orange rectangle, many pink sensors report traffic conditions of a huge residential area. So it is conceivable that they are grouped into one region. 
Furthermore, the red sensors are grouped into another region because they are in different directions of the same road, and are supposed to report different states. 
In the blue rectangle, the brown sensors are identified to be highly-related due to they belong to the same community. Besides, as the road goes to the mount area, the yellow sensors are recognized as another region.
These illustrations reflect the macro dependency captured of our regional graph, which utilizes the topology information and accords to the physical meaning.

In terms of the global perspective in Figure \ref{fig:METR_LA_Mog} (b), we can observe that sensors in the same color have POI of similar functions as neighbors.
There are consistently schools in the vicinity of the dark blue sensors, as shown in the magnified rectangles, reflecting the highly similar spatio-temporal patterns of these sensors. 
Also, the pink sensor always appears with recreational POIs, such as nature parks or theme parks, which implies that \name successfully identifies the function in the region where the sensor is located.
Notably, the mapping between regional and global graphs is soft, which means this observation may be a partial factor of the complex spatio-temporal patterns. 
However, we can still safely draw a conclusion that the common spatio-temporal patterns we mined can explain the properties of regions, which can provide auxiliary information for sensor modeling.  

\section{Related works}
\noindent\textbf{Spatio-Temporal Prediction.}
Spatio-temporal prediction involves predicting future trends by analyzing historical spatio-temporal properties.
Early approaches~\cite{ST-Res17,zhang2019flow,guo2019deep,li2020autost} were based on grid-based data, where the city is divided into grids, and Convolutional Neural Networks (CNN) are used to capture spatial dependencies between a grid and its neighbors. These approaches generally incorporate TCN~\cite{TCN} and RNN~\cite{RNN} to capture temporal variations.
For the later works~\cite{DCRNN-17, GWNET19, autostl} on graph-based data, 
Graph Neural Networks (GNN) are applied under the assumption that connected sensors in the graph have similar patterns within the same period. 

Basically, these approaches treat every sensor equally and overlook the implicit hierarchical structures present in the original graph.
With our \name, the inherent hierarchy can be used to enhance the representation of the original sensor graph.





\noindent\textbf{Hierarchical Spatio-Temporal Data Mining.}
Existing hierarchical spatio-temporal data mining works implement hierarchy construction by clustering node representations ~\cite{HGNN20,22HiGCN,22ST-HSL,hgcn21,MCSTGCN22,traffictmr}. Traffic-Transformer~\cite{traffictmr} and ST-HSL~\cite{22ST-HSL} capture global node dependencies using multi-head attention mechanisms and hypergraphs. 
HGNN~\cite{HGNN20} and HiGCN~\cite{22HiGCN} generate new hierarchies using unique trainable GAT or GCN models, while MC-STGCN \cite{MCSTGCN22} uses the Louvain algorithm~\cite{louvain} to detect potential communities. 

Some works~\cite{HGNN20,hgcn21,MCSTGCN22} attempt to build an implicit hierarchy in the graph's topological structure using spectral clustering~\cite{spectral01} but achieve suboptimal performance. Instead, our method efficiently solves the BCC to merge sensors with high intra-region dependency and generate the regional map with linear time complexity, which does not require updating or training and can be easily extended.
Other works~\cite{HGNN20,22HiGCN} construct unique graph neural networks for different hierarchies. In contrast, we incorporate a Meta GCN. By decoupling complex associations between sensors, we extract two hierarchical perspectives, 
which yields a comprehensive view of spatio-temporal representation learning.

\section{Conclusion}
The key to solving spatio-temporal prediction is how to model the dependency among all sensors.
Basically, existing models simply regard all nodes in an equal position, or devise a hierarchy based on simply clustering.
In this paper, we rethink the sensor's dependency modeling and propose our \name, to enhance spatio-temporal prediction with hierarchical information effectively.
We propose two hierarchical perspectives, \ie regional and global graphs, to describe the macro dependency among sensors and generate representative and common spatio-temporal patterns.
Furthermore, we elaborate the cross-hierarchy graph learning mechanism, \gcnname, to learn node representation in a meta-learning manner and facilitate information propagation among nodes in multiple levels. 



\begin{acks}
\vspace{-1mm}
    This research was partially supported by APRC - CityU New Research Initiatives (No.9610565, Start-up Grant for New Faculty of City University of Hong Kong), CityU - HKIDS Early Career Research Grant (No.9360163), Hong Kong ITC Innovation and Technology Fund Midstream Research Programme for Universities Project (No.ITS/034/22MS), SIRG - CityU Strategic Interdisciplinary Research Grant (No.7020046, No.7020074), SRG-Fd - CityU Strategic Research Grant (No.7005894), Tencent (CCF-Tencent Open Fund, Tencent Rhino-Bird Focused Research Fund), Huawei (Huawei Innovation Research Program), Ant Group (CCF-Ant Research Fund, Ant Group Research Fund) and Kuaishou.
    Hongwei Zhao is funded by the Provincial Science and Technology Innovation Special Fund Project of Jilin Province, grant number 20190302026GX, Natural Science Foundation of Jilin Province, grant number 20200201037JC, and the Fundamental Research Funds for the Central Universities, JLU.
    Zitao Liu is supported by National Key R\&D Program of China, under Grant No. 2022YFC3303600, and Key Laboratory of Smart Education of Guangdong Higher Education Institutes, Jinan University (2022LSYS003).
    
\end{acks}

\clearpage

\newpage
\bibliographystyle{ACM-Reference-Format}
\bibliography{9_Reference}

\end{document}